\documentclass[conference]{IEEEtran}
\usepackage{times}
\usepackage{graphicx}
\usepackage{picinpar}
\usepackage{amsmath}
\usepackage{amssymb}
\usepackage{amsfonts}
\usepackage{url}
\usepackage{flushend}
\usepackage[latin1]{inputenc}
\usepackage{colortbl}
\usepackage{soul}
\usepackage{multirow}
\usepackage{multicol}
\usepackage{booktabs}
\usepackage{pifont}
\usepackage{color}
\usepackage{alltt}
\usepackage{dsfont}
\usepackage[hidelinks]{hyperref}
\usepackage{enumerate}
\usepackage{siunitx}
\usepackage{epstopdf}
\usepackage{pbox}

\usepackage[numbers]{natbib}
\usepackage{multicol}

\def \textHT [#1]{\color{red}\textbf{#1}\color{black}}
\def \textLT [#1]{\color{blue}#1\color{black}}

\begin{document}

\title{i3dLoc: Image-to-range Cross-domain Localization Robust to Inconsistent Environmental Conditions}



%
\author{\authorblockN{Peng Yin\authorrefmark{1},
Lingyun Xu, Ji Zhang, Howie Choset and Sebastian Scherer}
\authorblockA{Robotics Institute, Carnegie Mellon University, Pittsburgh, PA 15213, USA.\\
Email: pyin2@andrew.cmu.edu, hitmaxtom@gmail.com, zhangji, basti@andrew.cmu.edu}
\authorblockA{\authorrefmark{1}Corresponding Author}}

\maketitle

\begin{abstract}
We present a method for localizing a single camera with respect to a point cloud map in indoor and outdoor scenes. 
The problem is challenging because correspondences of local invariant features are inconsistent across the domains between image and 3D. 
The problem is even more challenging as the method must handle various environmental conditions such as illumination, weather, and seasonal changes. 
Our method can match equirectangular images to the 3D range projections by extracting cross-domain symmetric place descriptors.
Our key insight is to retain condition-invariant 3D geometry features from limited data samples while eliminating the condition-related features by a designed Generative Adversarial Network.
Based on such features, we further design a spherical convolution network to learn viewpoint-invariant symmetric place descriptors.
We evaluate our method on extensive self-collected datasets, which involve \textit{Long-term} (variant appearance conditions), \textit{Large-scale} (up to $2km$ structure/unstructured environment), and \textit{Multistory} (four-floor confined space).
Our method surpasses other current state-of-the-arts by achieving around $3$ times higher place retrievals to inconsistent environments, and above $3$ times accuracy on online localization. 
To highlight our method's generalization capabilities, we also evaluate the recognition across different datasets. 
With a single trained model, i3dLoc can demonstrate reliable visual localization in random conditions.
\end{abstract}

\IEEEpeerreviewmaketitle

\section{Introduction}
Mobile robots and self-driving cars have entered our daily life in the recent years with the development of High-Definition maps-based accurate localization.
Cameras have the huge potential to provide low-cost, compact and self-contained visual localization against point cloud maps.
However, visual methods are inherently limited by inconsistent environmental conditions in the real world, e.g., illumination, weather, season and viewpoint differences.
Whereas, accurate matching can be challenging to perform on point cloud data due to sensor sparsity with no sufficient texture feature guarantees.
Transitional geometry-based methods~\cite{2D3D:Line-based} implicitly assume a static environment, such as stable lighting conditions, sunny weather, and fixed seasonal attributes.
Recent learning-based visual localization methods are either constrained under limit environments~\cite{2d3dmatchnet} (structure road) or only fit for limited viewpoints~\cite{vpr:dont} (forwards or backwards on the street).
Current image-to-range localization methods are difficult to leverage in real-world applications, or can hardly address the above issues simultaneously.

\begin{figure}[th]
    \begin{center}
        \includegraphics[width=\linewidth]{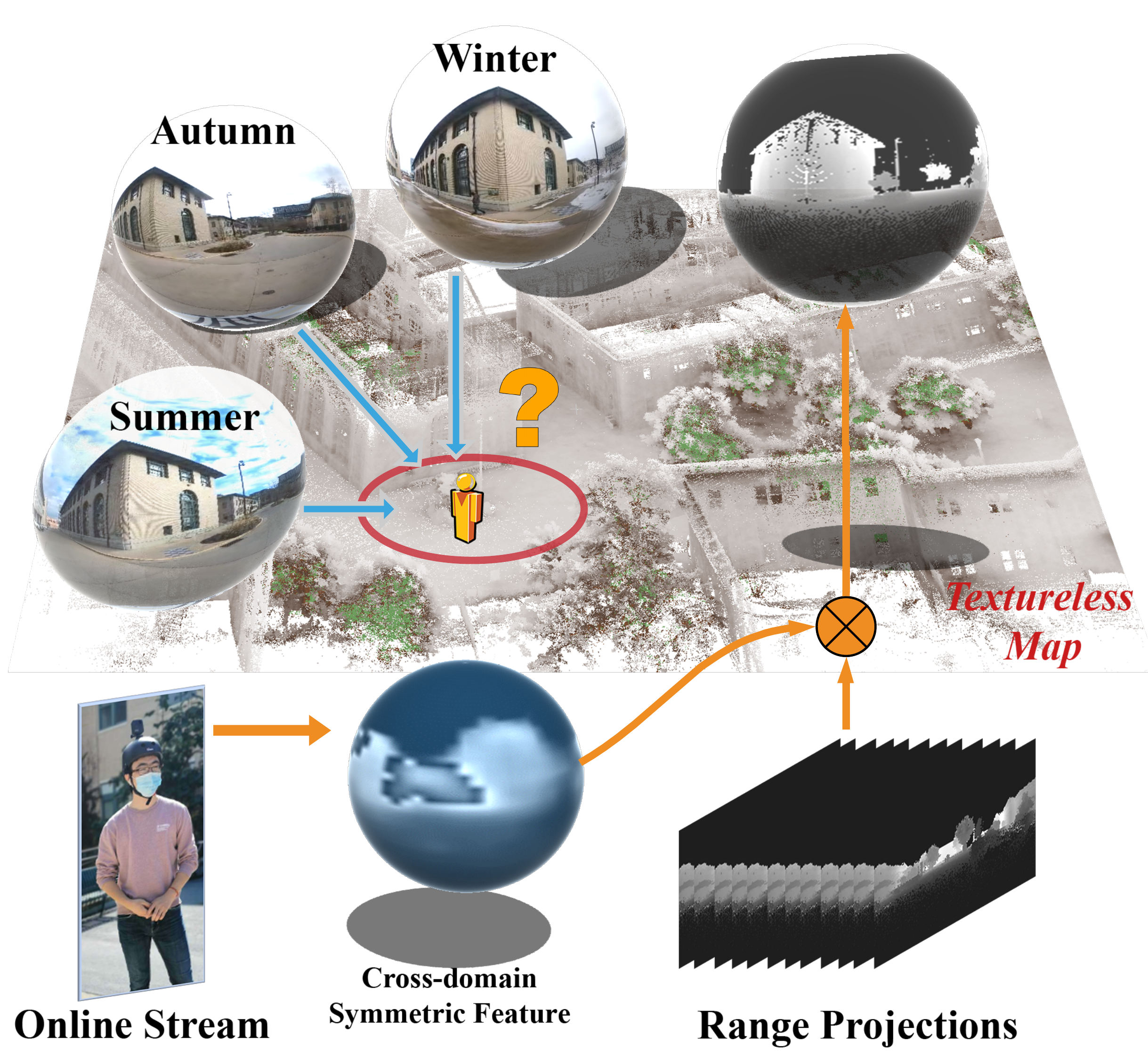}
    \end{center}
    \caption{
    Overview of our image-to-range cross-domain localization system.
    Under an inconsistent environment, we aim to localize a 360 camera within a prior textureless point cloud map.
    Given the 360 image and initial position, we extract potential range projections, which we then evaluate using symmetric feature distance for 3D position estimation.
    }
    \label{fig:idea}
\end{figure}

To fill this gap, we propose \textit{i3dLoc} -- a 3D visual localization method with the assistance of offline 3D maps -- invariant to environmental conditions and casual viewpoints.
Compared with Euclidean geometry features~\cite{2D3D:Line-based}, we exploit the symmetric place descriptors between equirectangular visual inputs and range projections under condition invariant feature domain.
As illustrated in Fig.~\ref{fig:idea}, i3dLoc tackles the 3D visual localization task through two steps:  
(1) retaining geometry features from raw visual inputs which eliminates condition-related (day/night lighting conditions) factors;
(2) extracting symmetric place descriptors even for inputs under vital viewpoint differences.


The major contributions of i3dLoc are:
\begin{itemize}
    \item We put forth a new end-to-end large-scale visual localization method with the assistance of offline 3D maps, providing reliable 3D localization.
    \item
    We introduce a Generative adversarial Network (GAN) based cross-domain transfer learning network to extract condition-invariant features 
    while eliminating the condition-related factors.
    \item
    We design an innovative symmetric feature learning network performing on spherical convolution networks, where intrinsic characteristics of a spherical harmonica naturally help place descriptor matching under variant viewpoints.
    \item
    We design an evaluation framework (includes \textit{Long-term}, \textit{Large-scale} and \textit{Multistory} datasets), which can analyze the visual localization performance under significant environmental appearance changes, casual viewpoints and also the generalization ability for unseen datasets.
\end{itemize}

In summary, i3dLoc provides a low-cost, condition- and viewpoint-invariant visual localization method for both indoor and outdoor large-scale environments.
As demonstrated in the experimental results, i3dLoc outperforms the state-of-the-art image-to-range visual localization methods~\cite{2d3dmatchnet} and significantly improves the localization accuracy of traditional visual SLAM methods~\cite{VPR:orbslam3}.
When trained with all self-collected datasets under variant conditions, our method also shows great generalization ability for unseen indoor and outdoor environments, which makes it suitable for low-cost but robust localization for mobile robots.

\section{Related Works}
\label{sec:related_works}
Visual localization has been well studied in the traditional SLAM framework; we refer to Lowry \emph{et.al}~\cite{VPR:SURVEY} for an overview of approaches using cameras.
The 3D point cloud based localization methods have also been well addressed~\cite{PR:pointnetvlad,PR:LPDNet,PR:overlapnet} recently.
Image-to-range localization based on consistent 3D maps and low-cost camera sensors have attracted more and more attention in recent years.
Here we mainly concentrate on the related image-to-range localization approaches.

There are two main trends in image-to-range localization: the geometry based feature matching methods~\cite{2D3D:corner, 2D3D:Line-based} and data-driven based visual localization methods~\cite{PR:netvlad,2d3dmatchnet}.
Geometry features are usually carefully designed to bridge the description gap between image and point cloud domains and maintain the geometry's consistent nature.
Xie \emph{et.al}~\cite{2D3D:corner} propose to extract corner points of known calibration targets to maintain image-to-range feature connections.
Compared with point features, line features in structured environments are more consistent for both image and range projection domains. Yu \emph{et.al}~\cite{2D3D:Line-based} introduce an image-to-range registration method based on 2D and 3D line correspondences for place recognition.

Recent approaches have leveraged deep learning to develop data-driven visual localization frameworks that outperform classical methods both in accuracy and speed.
In~\cite{2d3dmatchnet}, Feng~\emph{et.al} propose an end-to-end deep network architecture to jointly learn the descriptors for 2D and 3D keypoints from images and point clouds.
This method learns the cross-domain features through a weighted soft-margin triplet loss, while ignoring the underlying geometry connections between the two different domains.
Sun~\emph{et.al}~\cite{2D3D:DepthReg} introduce an image-to-range coarse localization method by building the feature connections within depth images, where the depth is estimated from a depth prediction network.

Contrary to the methods mentioned above, our method, i3dLoc exploits the more general image-to-range visual localization solution.
Similar to~\cite{2d3dmatchnet}, we utilize a triplet-like loss to aggregate the learned 2D and 3D features into global descriptors.
The major difference is that our method does not restrict the viewpoints and environmental conditions of visual inputs, which improves the generalization ability in real applications.
Our method also show high robustness to unstructured environments, where corresponding features between image and range projections are hard to find.

\section{Methodology}
\label{sec:method}
    As illustrated in Fig.~\ref{fig:idea}, the main idea of i3dLoc is to find the corresponding place descriptors from 2D images and 3D range projections that are invariant to environmental conditions and viewpoints.
    To deal with the effects from appearance changes and viewpoint differences, i3dLoc mainly includes two modules:
    (1) a cross-domain transfer network to transform 2D equirectangular images into condition-invariant range projections; 
    (2) a symmetric feature learning network to extract viewpoint-invariant descriptors even under casual viewpoints.  
    To enable the end-to-end training, we design the domain transfer metric to enhance the domain adaptation, and the triplet-like learning metrics to bridge the feature similarity between 2D images and 3D range projections.
    
    
    \begin{figure*}[!th]
        \centering
        \includegraphics[width=\linewidth]{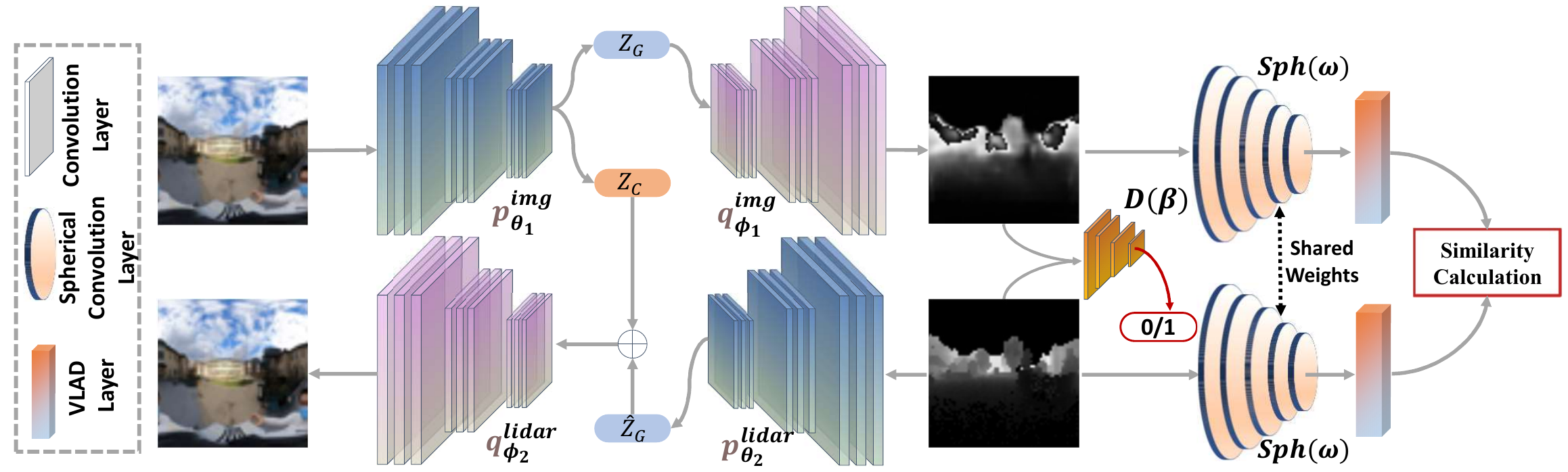}
        \caption{The framework of the proposed i3dLoc.
        It is the combination of two generator networks between 2D imagery domain and range prediction domain, a classification module estimating the environmental conditions, and a discriminator module distinguishing the generated range predictions from the real LiDAR projections.}
        \label{fig:framework}
    \end{figure*}

    \subsection{Cross-domain Transfer Learning}
    \label{sec:DTM}
    To generate constant geometry features from visual inputs under different environmental conditions, we construct a cross domain transfer network between 2D imagery and 3D range projections.
    Before we introduce the details, we will introduce the visual features from the point view of information entropy.
    Naturally, the condition- related feature $Z_{C}$ (illumination, weather, seasons) and invariant features $Z_{G}$ (geometry) in the image domain are tightly coupled.
    $H(Z_{G}, Z_{C}|x)$ and $I(Z_{G};Z_{C}|x)$ are the joint entropy and the mutual entropies conditioned on the given data samples $x$ in the image domain.
    $H(Z_{G}|Z_{C},x)$ and $H(Z_{C}|Z_{G},x)$ are the conditional entropies based on $Z_{C}$ and $Z_{G}$ respectively.

    In the visual localization task, gathering visual data samples under all kinds of environmental conditions for each single area is difficult and time-consuming.  
    To learn condition-invariant place features with limited data samples, we mainly focus on the following three aspects:
    decreasing the joint entropy $H(Z_G, Z_C|x)$, this approach can reduce the uncertainty between visual inputs and corresponding place features;
    improving the conditional entropy $H(Z_G|Z_C,x)$, this can enhance place features capturing more non-geometry structures $Z_G$ from limited samples;
    reducing the mutual entropy $I(Z_{G};Z_{C}|x)$, this can help separate the geometry features $Z_{G}$ from the non-geometry features $Z_{C}$ within the same visual distribution.

    The joint entropy $H_{p_{\theta_1}^{img}}(z|x), z\in\{Z_G,Z_C\}$ measures the uncertainty of extracted geometry and non-geometry features with the given data samples $x$.
    Thus, reducing $H_{p_{\theta_1}^{img}}(z|x)$ can improve the uniqueness mapping from $x$ to $z$, where $p_{\theta_1}^{img}$ is the parameter in the encoder module from the raw image to range predictions as depicted in Fig.~\ref{fig:framework}. 
    However, improving the conditional entropy $H_{p_{\theta_1}^{img}}(z|x)$ is intractable, since we can not access the data-label pair between the visual inputs $x$ and their corresponding features $\{Z_G, Z_C\}$.
    Especially in the localization task, where each area has variants of visual appearances, i.e., there exists different kinds of combinations between $Z_G$ and $Z_C$.
    An alternative approach is to optimize the upper bound of $H_{p_{\theta_1}^{img}}(z|x)$ through,
    \begin{align}
        &\min_{\theta_1}H_{p_{\theta_1}^{img}}(z|x) \label{ep:joint_entropy} \\
        &\triangleq \min_{\theta_1} -\sum p_{\theta_1}^{img}(z|x)\log(p_{\theta_1}^{img}(z|x))  \nonumber \\
        &=\min_{\theta_1, F} -\sum p_{\theta_1}^{img}(z|x)[\log(F(z|x))]   \nonumber
        \\
        &    -\sum p_{\theta_1, F}^{img}(z|x)[\log(p_{\theta_1}^{img}(z|x))-\log(F(z|x))]  \nonumber 
        \\
        &=\min_{\theta_1, F}  H_{p_{\theta_1}^{img}(z|x)}[\log(F(z|x))] \nonumber \\
        &    -E_{p_{\theta_1}^{img}(z|x)}[\mathbf{KL}(p_{\theta_1}^{img}(z|x)\|(F(z|x)))]
        \nonumber
    \end{align}
    where $\mathbf{KL}$ is Kullback-Leibler divergence, which measures the distance of two data distributions.
    $H_{p_{\theta_1}^{img}}(\log(F(z|x)))$ measures the uncertainty of the predicted place feature with a given sample data $x$.
    $F(\cdot)$ is the combination modules of $q_{\phi_1}^{img}$ and $p_{\theta_2}^{lidar}$ as depicted in Fig.~\ref{fig:framework}, which can reconstruct the geometry features based on the original $Z_G$.
    Since $\mathbf{KL}(\cdot)\geq 0$, Eq.~\ref{ep:joint_entropy} can be rewritten as,
    \begin{align}
        &\min_{\theta_1} H_{p_{\theta_1}^{img}}(z|x) \leq \min_{\theta_1, F} H_{p_{\theta_1}^{img}(z|x)}[\log(F(z|x))] \label{ep:loss_recon} \\
        &\triangleq \min_{\theta_1, F, \phi_2} H_{x}[\log(q_{\phi_2}^{lidar}(F(z|x)))] \nonumber \\
        &=\min_{\theta_1, \phi_1, \theta_2, \phi_2} H_{\hat{x}\sim  \{p_{\theta_1}^{img},q_{\phi_1}^{img},p_{\theta_2}^{lidar},q_{\phi_2}^{lidar}|x\}}[\log(x=\hat{x}|x))] \nonumber \\
        &=\mathcal{L}_{Recon}(x, \hat{x}) \nonumber 
    \end{align}
    where $\mathcal{L}_{Recon}$ is the reconstruction loss between the original visual input $x$ and the reconstructed image $\hat{x}$.
    The original $H_{p_{\theta_1}^{img}}(z|x)$ is transformed into its upper bound $\mathcal{L}_{Recon}(z, \hat{z})$.

       \begin{figure}[t]
        \centering
        \includegraphics[width=\linewidth]{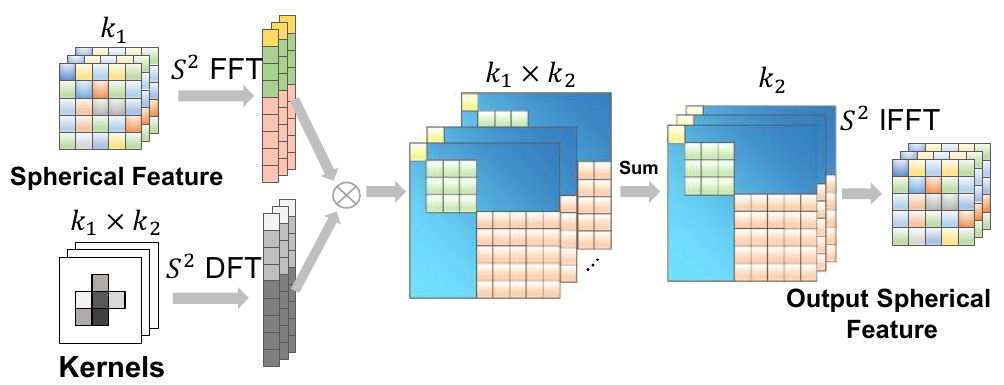}
        \caption{The spherical convolution module.
        With the given spherical feature $f$ and a kernel signal $h$, we first transform them into the harmonic domain ($H_f$, $H_h$)with the Fast Fourier transform (FFT) and Discrete Fourier Transform (DFT) respectively.}
        \label{fig:sphnet}
    \end{figure}

    To improve the conditional entropy $H(Z_G|Z_C,x)$, we design a generative adversarial network (GAN)~\cite{CNN:GAN} for the extracted LiDAR projections and visual range predictions,
    \begin{align}
        &\mathcal{L}_{GAN} 
        = \min_{\theta_1, \phi_1}\max_{\beta}E(\log(D_{\beta}(y)) +\label{ep:loss_gan} \\
        &E_{\{z_G,z_C\}\sim p_{\theta_1}^{img}(z|x),\hat{y}\sim q_{\phi_1}^{img}(y|z_G)}(\log(1-D_{\beta}(\hat{y})))\nonumber
    \end{align}
    where $y$ in the first term is the real LiDAR projection, and $\hat{y}$ in the second term is the generated range predictions based on the estimated geometry features $z_G$ from the visual input $x$.
    As demonstrated by Goodfellow \emph{et.al}~\cite{CNN:GAN}, with iterative updating of the decoder and the discriminator module, GAN can pull the data distribution of generated data closer to the target data, and meanwhile improve the decoder's generalization ability.
    
    Finally, the mutual entropy $I(z_{G};z_{C}|x)$ can be extended by,
    \begin{align}
        I(z_{G};z_{C}|x) = H(z_{G}|x)-H(z_{G}|z_{C},x)\label{eq:mutual_entropy}
    \end{align}
    where, reducing the mutual entropy is equivalent to reducing the right-hand term in the above equation.
    We provide the paired data samples of visual images $x$ and corresponding range projections $y$; both data can be extracted under the same viewpoint and $y$ only contains the geometry features.
    The geometry feature distribution $H(Z_G|x)$ can be estimated from the encoder module $\{z_G, z_C\}\sim p_{\theta_1}^{img}(z|x)$, and the ideal conditional geometry feature distribution is equal to $\hat{z}_G \sim p_{\theta_2}^{lidar}(z|y)$.
    We apply a soft triplet loss to reduce the feature difference between $z_G$ and $\hat{z}_G$,
    \begin{align}
        L_{M}(\mathcal{T}) = \lambda_1 + d(Z_{G}, \hat{Z}_G) - d(Z_{G}, Z_{C}) \label{ep:loss_mutual}
    \end{align}
    $d(\cdot)$ denotes the Euclidean distance and $\lambda_1$ is the hyper-parameter to control the feature distance.
    Based on Eq.~\ref{ep:loss_recon}, \ref{ep:loss_gan} and \ref{ep:loss_mutual}, we can construct the conditional domain transfer module from visual inputs to range projections,
    \begin{align}
    \mathcal{L}_{Transfer} = \mathcal{L}_{Recon}+\mathcal{L}_{GAN}+\mathcal{L}_{M}(\mathcal{T})
    \label{eq:domain_transfer}
    \end{align}
    
    \subsection{Symmetric Feature Learning}
    \label{sec:Feature_learning}
    To learn viewpoint-invariant descriptors for both range predictions from visual inputs and range projections from 3D maps, we utilize the spherical harmonics to learn the place descriptors.
    As illustrated in Fig.~\ref{fig:idea}, representing data in the spherical view is natural in 3D geometry analysis; however, it is difficult to adopt traditional convolution network in the spherical domain, because spaces between adjacent points in the spherical view are not uniform.
    Instead of traditional convolution, we apply the spherical convolution based on the property of spherical harmonics.
    Spherical convolution avoids space-varying distortions in the Euclidean space by convolving spherical signals in the harmonic domain.
    The mathematical model of spherical convolution into the harmonic domain shows its orientation-equivalent.
    Spherical convolution of $SO(3)$ signals $f$ and $h$ ($f, h$ are functions: $SO(3) \rightarrow \mathbb{R}^K$) in the rotation group $SO(3)$ are defined as,
    \begin{align}
        [f \star_{SO(3)} h](\mathbf{R}) = & \int_{SO(3)}f(\mathbf{R}^{-1}\mathbf{Q})h(\mathbf{Q})d\mathbf{Q}
    \end{align}
    where $\mathbf{R, Q} \in SO(3)$. 
    As the proof in~\cite{cohen2018spherical}, spherical convolution is shown to be orientation-equivariant,
    \begin{align}
        [f \star_{SO(3)} [L_{Q}h](\mathbf{R})
        = & [L_{\mathbf{Q}}[f \star_{SO(3)} h]](\mathbf{R})          
    \end{align}
    where $L_{\mathbf{Q}} (\mathbf{Q} \in SO(3))$ is a rotation operator for spherical signals.
    As depicted in Fig.~\ref{fig:sphnet}, the convolution of two spherical signals $f$ and $h$ in the harmonics domain are computed by three steps.
    We first expand $f$ and $h$ to their spherical harmonic basis $H_f$ and $H_h$, then compute the point-wise product of harmonic coefficients, and finally invert the spherical harmonic expansion.
    For more details, we suggest the reader refer to the original work in~\cite{Sphere:SO3_invariant}.
    
    Intuitively, there exists spatial similarity in local outputs of spherical convolution.
    To leverage the viewpoint-invariant feature extraction, we utilize the VLAD layer~\cite{PR:netvlad}, which can cluster the local features into the global place descriptors.
    With the assistance of our cross-domain transfer module, we can learn the conditional- and viewpoint-invariant place descriptors from visual inputs to match the static range projections.

    \subsection{Learning Metrics}
    \label{sec:learn_metric}
    To enable the end-to-end training for visual localization, we introduce triplet-like learning metrics to learn conditional- and viewpoint-invariant place descriptors.
    For the convenience of illustrating loss functions, we first describe the necessary definitions. 
    The training tuple in both visual/LiDAR domains consists of four components:
    $\mathcal{S} = [S_a, \{S_{rot}\}, \{S_{pos}\}, \{S_{neg}\}\}]$, where $S_a$ is the spherical projections at the given position.
    $\{S_{rot}\}$ is a set of spherical representations manually rotated from $\{S_a\}$, where the rotation angles are random sampled from  ($[\ang{0}, \ang{30},...\ang{330}]$).
    $\{S_{pos}\}$ denotes a set of spherical representations of 3D scans (``positive'') whose distance to $\{S_a\}$ is within the threshold $D_{pos}$, and $\{S_{neg}\}$ denotes a set of 3D scans (``negative'') whose distance to $\{S_a\}$ is beyond $D_{neg}$.
    In our applications, we set the threshold $D_{pos}=5m$ and $D_{neg}=20m$.
    We construct paired tuples within visual domain $\mathcal{S}^{V}$ and 3D map domain $\mathcal{S}^{L}$.
    Ideally, we want to minimize feature distances in both domains: 
    \begin{align}
        &L_{View}(\mathcal{T}) = \\
        &\max_{i, j}([\lambda_2 + d(f(S_a), f(S_{pos_i})) -
        d(f(S_a), f(S_{neg_i}))]_+) + \nonumber \\
        &\max_{i,j,k}([\lambda_3 + d(f(S_{rot_j}), f(S_{pos_i})) -d(f(S_{rot_j}), f(S_{neg_i}))]_+) \nonumber
    \end{align}
    $f(.)$ is the function that encodes spherical representations into global descriptors by symmetric feature learning module, and $d(\cdot)$ denotes the Euclidean distance.
    $[.]_+$ denotes the hinge loss, $\lambda_2$ and $\lambda_3$ are the constant thresholds to control the margins between the feature differences of different Euclidean distances.
    Meanwhile, we also define a domain learning metric to reduce the cross-domain feature differences:
    \begin{align}
        &L_{Domain}(\mathcal{T}) = \\
        &\max_{i, j}([\lambda_4 + d(f(S_a^{V}), f(S_{pos_i}^{L})) -
        d(f(S_a^{V}), f(S_{neg_i}^{L}))]_+) + \nonumber \\
        &\max_{i,j,k}([\lambda_5 + d(f(S_{rot_j}^{V}), f(S_{pos_i}^{L})) -d(f(S_{rot_j}^{V}), f(S_{neg_i}^{L}))]_+) \nonumber
    \end{align}
    $\lambda_4$ and $\lambda_5$ are the constant thresholds to control the margins between the feature differences under visual/LiDAR domains.
    By combining the domain transfer metric $\mathcal{L}_{Transfer}$ and the above place learning metrics, the final joint learning metric can be written as,
    \begin{align}
        \mathcal{L}_{Joint} = \mathcal{L}_{Transfer} +\mathcal{L}_{Domain}(\mathcal{T})+L_{View}(\mathcal{T}) \nonumber \label{ep:joint}
    \end{align}
    In our application, $\lambda_1$, $\lambda_2$ and $\lambda_4$ is set to $0.5m$ and $\lambda_3$ and $\lambda_5$ is set to $1.0m$.
    During the training procedure, we first train the domain transfer module with paired images and LiDAR projections; then we use the pertained transfer model for the conditional- and viewpoint-invariant place descriptors.

\section{Experiments}
\label{sec:experiments}
\begin{figure}[t]
        \begin{center}
            \includegraphics[width=0.8\linewidth]{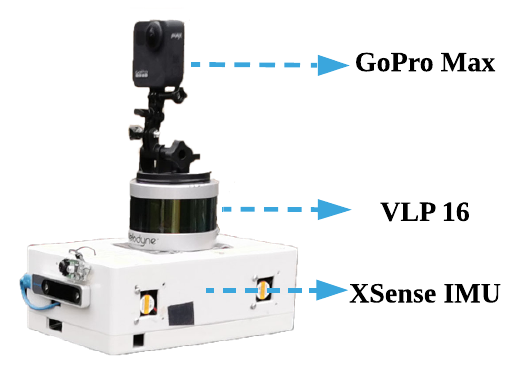}
        \end{center}
        \caption{
        The data-collection platform.
        This platform includes a LiDAR device, an 360 camera and an inertial measurement unit (IMU).
        We gathered the indoor and outdoor datasets via recording the raw data from above sensors.
        Since lacking the ground truth position in indoor or other GPS-denied environments, we use the LiDAR odometry outputs~\cite{LOAM:zhang2014loam} as the ground truth estimation.
        }
        \label{fig:platform}
    \end{figure}

    In this section, we demonstrate the visual localization performance of \textit{i3dLoc} on both indoor and outdoor datasets generated by our data collection platform as depicted in Fig.~\ref{fig:platform}.
    In all experiments, we utilize a LiDAR device (Velodyne-VLP 16) and an omnidirectional camera (GoPro Max) mounting on the top of the payload, an inertial measurement unit (Xsense MTI $30$, $0.5^{\circ}$ error in roll/pitch, $1^{\circ}$ error in yaw, $550m$W), a mini PC (Intel NUC i7, $3.5$ GHz, $28$W) and an embedded GPU device (Nvidia Xavier, $8$G memory).
    The network is trained on a GPU server with a single Nvidia 1080Ti GPU and $64$G RAM.
    In the rest of this section, we detail the datasets, comparison methods, and evaluation metrics respectively.
    Then, we analyze qualitatively and quantitatively the performance of \textit{i3dLoc} on place retrieval and online localization.
    Finally, we further discuss the current limitations and failure cases of i3dLoc.

    \bigskip
    \noindent \textbf{Datasets}.
    The training and evaluation dataset includes:
    \begin{itemize}
        \item \textbf{Long-term Dataset}, we create $15$ long-term dataset by generating trajectories in $2020/06\sim2021/01$ with variant season, weather changes. 
        Distance for each trajectory is around $200\sim 400m$. 
        Trajectories $\{1\sim 10\}$ are fed into the training procedure, and $\{11\sim 15\}$ for evaluation.
        \item \textbf{Large-scale Dataset}, we create a large-scale outdoor dataset with $8$ trajectories by traversing $1.5\sim 2km$ routes under structured/unstructured outdoor environments. 
        Trajectories $\{1\sim 6\}$ are fed into the training procedure, and trajectories $\{7\sim 8\}$ are for evaluation.
        \item \textbf{Multistory Dataset}, we create a indoor dataset by traversing $8$ trajectories within a multi-floor area under daytime/nighttime.
        The average distance for indoor routines is $100\sim 150m$.
        We use trajectories $\{1\sim 6\}$ for network training, and $\{7\sim 8\}$ for evaluation.
    \end{itemize}

    
    \begin{table}[ht]
        \centering
        \caption{Dataset frames splitting in training/evaluation.}
        \begin{tabular}{c c c c}
        \toprule
        & \textit{Long-term} & \textit{Large-scale} & \textit{Multistory}\\ \midrule
        Train            & $13,070$         & $15,971$      & $13,830$  \\
        Evaluation       & $3,268$          & $3,993$       & $3,458$   \\
        Distance         & $200\sim 400m$   & $1.5\sim 2km$ & $100\sim 150m$    \\
        \bottomrule
        \label{table:dataset}
        \end{tabular}
    \end{table}

    All the above datasets are collected by simultaneously holding the data-collection platform and recording LiDAR, IMU, and 360 images.
    To provide training/evaluation data, we first generate the global map with all the LiDAR sequences through a traditional LiDAR odometry method~\cite{LOAM:zhang2014loam}.
    Since we are working in indoor and outdoor GPS-denied environments, we can not obtain ground truth position from a third-party system. 
    In this paper, we use the LiDAR odometry estimation as the ground truth.
    Also, because the 3D offline map is generated using the same LiDAR odometry method, the standard division of relative ground truth noise is tiny.
    Based on this ground truth, we generate the paired images and 3D range projections.
    The range projections are generated by projecting points within $30m$ back to the keyframe.
    We resize the 2D images and 3D range projections to $64\times64$. 
    Table.~\ref{table:dataset} shows the data splitting in the training and evaluation procedure into three different datasets.
    We evaluate the condition-invariant property on the \textit{Long-term} dataset, which includes a different combination of environmental conditions.
    To investigate the viewpoint-invariant property, we generate the visual images on the same trajectory of different datasets but with casual viewpoint differences. And we also analyze the visual localization results when trained with non-rotated datasets but infer with the rotated datasets.
    Finally, we investigate the generalization ability by training one single model on the above three datasets and infer on unseen indoor and outdoor environments.

    \bigskip
    \noindent \textbf{Evaluation Metrics and Methods}
    We first consider the place retrieval performance under changing conditions and viewpoints by comparing our method with several baseline methods, NetVLAD\footnote{https://github.com/Nanne/pytorch-NetVlad}~\cite{PR:netvlad}, and 2D3DMatchNet~\cite{2d3dmatchnet}.
    We also combine our domain transfer module and NetVLAD, named i3d-Net, to compare with i3dLoc under viewpoint differences.
    For the 2D3DMatchNet, we manually imply their method for only position estimation without orientation estimation.
    Since all the above methods are image retrieval approaches that approximate the query's pose, we use the Average Recall at top $1\%$ retrievals. The threshold for success retrieval distance is set to ($10m$).
    Please notes our localization evaluation does include position, but not orientation.
    All the above learning-based methods are trained under the same data configuration.
    
    Secondly, we combine the visual localization and odometry for global localization.
    We investigate the online localization performance on the \textit{Large-scale} dataset by evaluating the absolute pose error (APE) with Evaluation of Odometry tool~\footnote{https://github.com/MichaelGrupp/evo}.
    We also compare with the pure visual odometry approach, ORB-SLAM\footnote{https://github.com/UZ-SLAMLab/ORB\_SLAM3}~\cite{VPR:orbslam3}.

 \begin{figure*}[ht]
        \centering
        {{\includegraphics[width=0.45\linewidth]{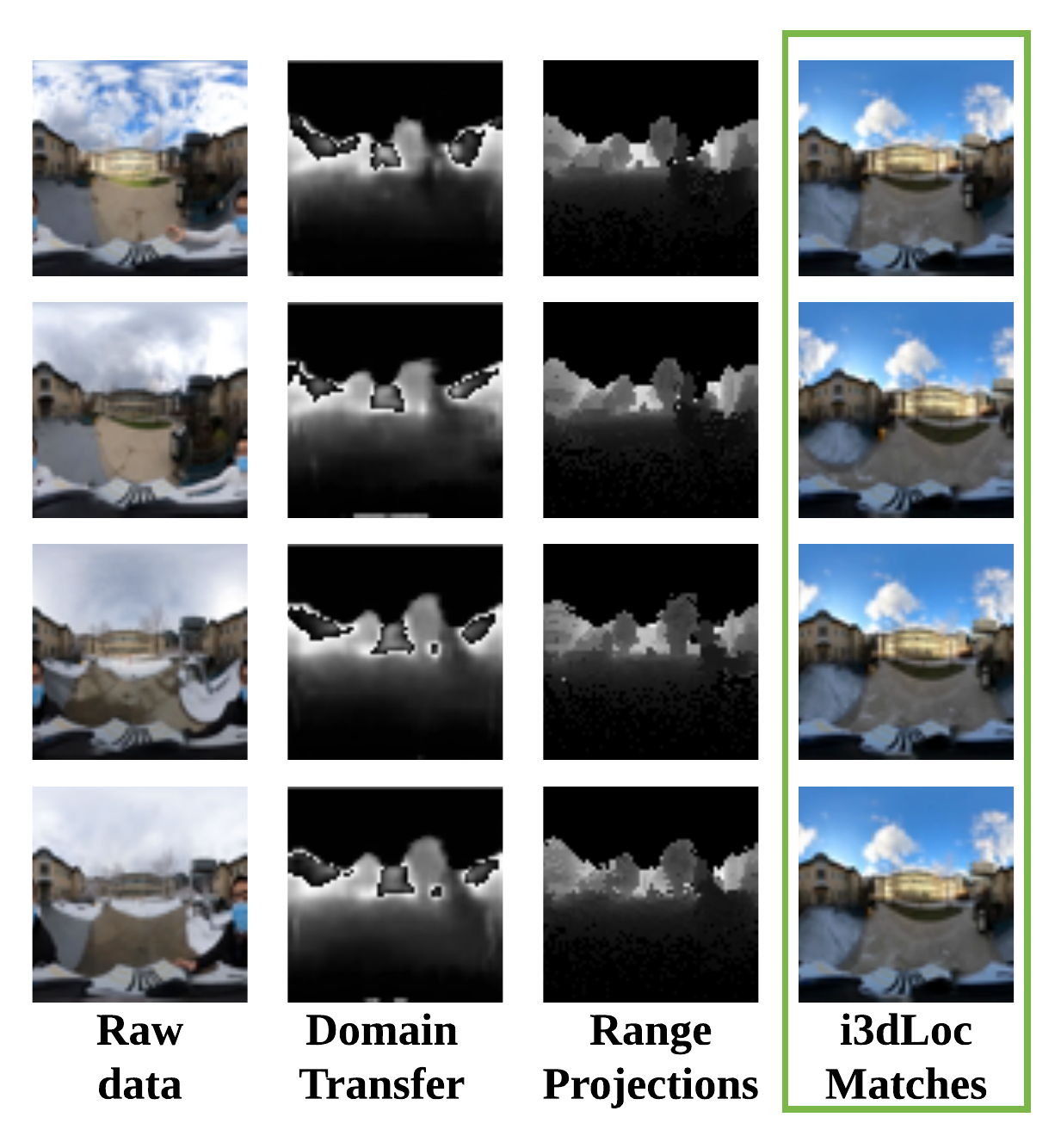} }}
        {{\includegraphics[width=0.45\linewidth]{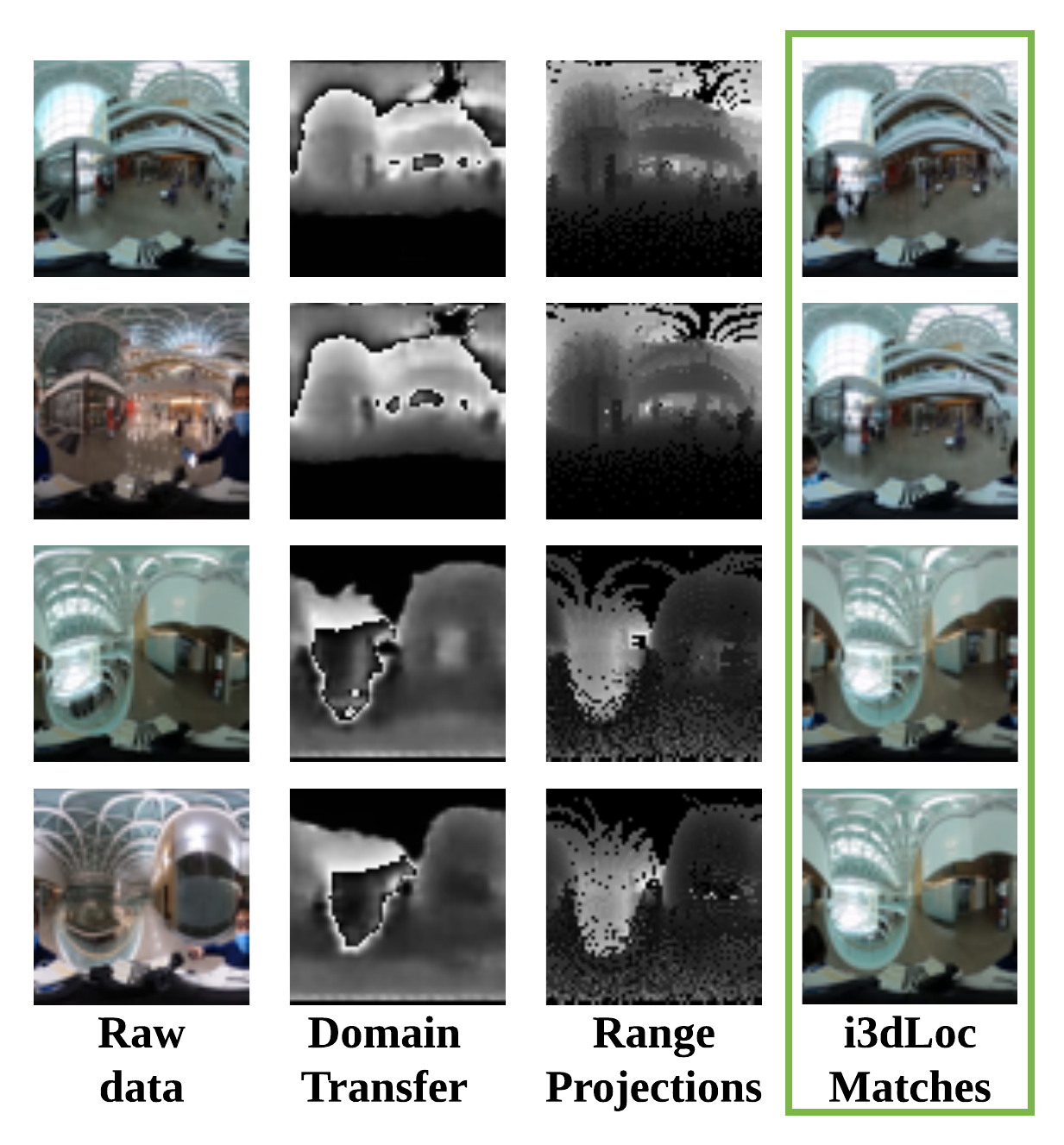} }}%
        \caption{Condition-invariant outdoor recognition.
        The first column shows the raw images under different conditions, the second column shows estimated range images with the domain-transfer module, last column shows the matched range projections of i3dLoc}
    \label{fig:domain_transfer}%
    \end{figure*}

\subsection{Place Retrieval Results}
\label{sec:PR_results}
\subsubsection{Condition-invariant Analysis}
\label{sec:condition_analysis}
We investigate condition-invariant property on the \textit{Long-term} and \textit{Multistory} dataset.
For fair comparison, all training/testing datasets are fed with fixed viewpoints.
Fig.~\ref{fig:domain_transfer} shows the matching results of \textit{Long-term} and \textit{Multistory} datasets.
The first column shows the raw equirectangular images from the 360 camera, and each image is taken from different lighting, season and weather conditions.
Given the visual inputs, we analyze the matching results on fixed day-time conditions of different learning-based methods.
The second column shows estimated geometry predictions by i3dLoc with the same domain-transfer module, where we can notice that the predictions reduce the condition factors but keep the geometry outlines.
The third column shows the range projections, which is generated by projecting the surrounding point cloud onto spherical view with the relative keyframe on the 3D map.
We can see that the estimated range projections from different conditions share similiar geometric structures for both indoor and outdoor environments.
The last three columns show the matching results of i3dLoc, NetVLAD and 2D3DMatchNet respectively.
Compared to other methods, i3dLoc can provide reliable visual retrieval under variant environmental conditions. 
    
    \begin{table}[t]
        \centering
        \caption{
        The average recall of  top $@1$ of different datasets.
        \textit{L+M} means the combination of \textit{Long-term} and \textit{Multistory} datasets}
        \begin{tabular}{c c c c c c}
        \toprule
        Method & \textit{Long-term} & \textit{Multistory} & \textit{L+M} \\ \midrule
        NetVLAD~\cite{PR:netvlad}        & $27.82\%$ & $21.24\%$ & $18.19\%$ \\ 
        2D3DMatchNet~\cite{2d3dmatchnet} & $23.34\%$ & $15.32\%$ & $8.92\%$ \\ 
        i3dLoc ($\mathcal{L}_{reco}$)  & $72.33\%$ & $33.34\%$ & $41.31\%$\\
        i3dLoc ($\mathcal{L}_{reco}+\mathcal{L}_{G}$) & $73.41\%$ & $35.12\%$ & $42.55\%$\\
        i3dLoc ($\mathcal{L}_{reco}+\mathcal{L}_{M}$) & $76.54\%$ & $38.25\%$ & $43.91\%$\\
        i3dLoc & $87.14\%$ & $42.62\%$ & $47.26\%$\\
        \bottomrule
        \label{table:pr_cond}
        \end{tabular}
    \end{table}

    \begin{figure*}[ht]
        \begin{center}
        \includegraphics[width=\linewidth]{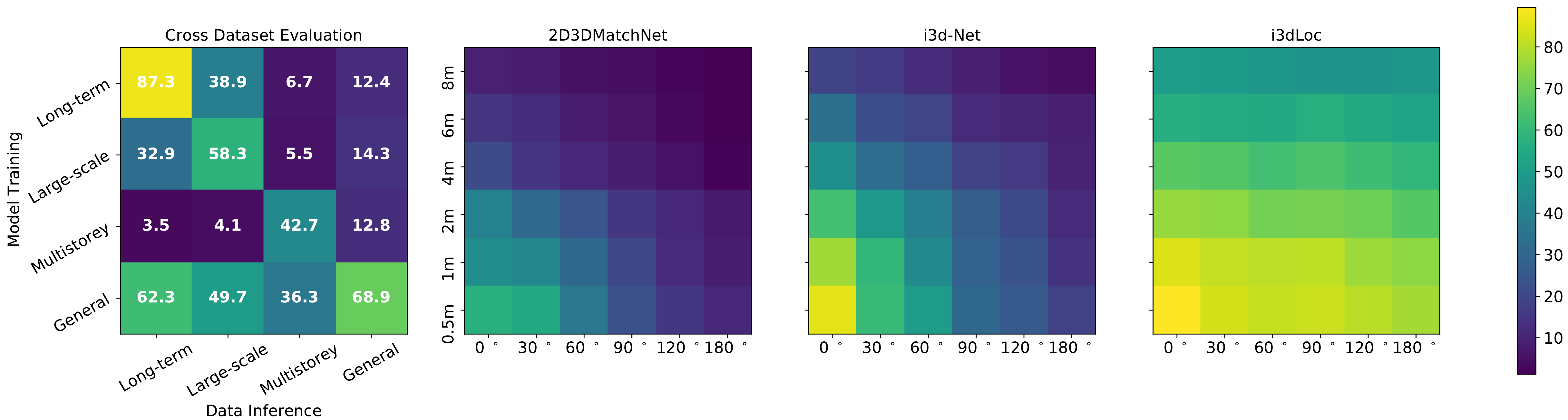}
        \end{center}
        \caption{Average recall of the top $1\%$ retrievals between image and range data.
        In the left figure, we train and evaluate i3dLoc under different datasets.
        Here \textit{General} dataset is the combination of all the three collected datasets.
        The right three figures show place retrievals under different viewpoints on the \textit{Large-scale} dataset. 
        We introduce viewpoint difference by manually adding translation and orientation in the range projections.}
        \label{fig:recall}
    \end{figure*}

Table.~\ref{table:pr_cond} gives the quantitative analysis of average recall at top $1\%$ retrievals of the different learning-based methods.
To evaluate the generalization ability, except \textit{Long-term} and \textit{Multistory} datasets, we also analyze the visual localization results on the \textit{General} datasets, which is the combination of \textit{Long-term} and \textit{Multistory} datasets, but with only $50\%$ original training data. 
To further investigate the effects of our proposed domain transfer, we also compare the place retrieval performance under different combinations of loss metrics.
The performance of i3dLoc ($\mathcal{L}_{reco}$) has outperformed other learning-based baselines, and we can note that environmental conditions has significant effect on the place retrieval accuracy.
However, only with reconstruct module has very limited improvements on the \textit{General} dataset, which indicates that the learned domain transfer module has limited generalization ability on unseen datasets.
When we combine $\mathcal{L}_{reco}$ with the GAN module $\mathcal{L}_{G}$ or with the mutual information module $\mathcal{L}_{M}$, the performance on \textit{General} dataset can be further improved.
Finally, with the complete cross-domain transfer module i3dLoc can outperform other learning-based methods on environments with complex environmental conditions, and also has higher generalization ability for unseen datasets.

\subsubsection{Viewpoint-invariant Analysis}
\label{sec:viewpoint_analysis}
    To evaluate the visual localization accuracy under variant viewpoints, we analyze the top $1\%$ retrievals on different datasets with rotated datasets.
    Given the same trajectory, we provide the same visual inputs, but with range projections with different viewpoints.
    Since range projections are generated by projecting surrounding points onto the trajectory way points, thus we can generated different range projections by manually adding translation and orientation to the waypoints.
    As shown in the right three figures of Fig.~\ref{fig:recall}, the y-axis represents the translation difference (from $0\sim 10$m), x-axis represents the orientation difference (from $0\sim180^{\circ}$).
    Here, we mainly show the comparison results with i3d-Net and 2D3DMatchNet.
    i3d-Net shows higher viewpoint-invariant property than 2D3DMatchNet, since the VLAD layer can extract an order-invariant place descriptor from local features.
    i3dLoc shows even higher robustness to all viewpoint differences, especially for orientations.
    Before being sent to the VLAD layer, the extracted features from the symmetric learning module are orientation equivalent.
    
    \begin{table}[t]
        \centering
        \caption{The average recall of top $1\%$ on three datasets.
        Tr-R/NR: training on rotated/non-rotated dataset, Te-R: testing on rotated dataset.}
        \begin{tabular}{c c c c c c}
        \toprule
        Method & \textit{Long-term} & \textit{Large-scale} & \textit{Multistory} \\ \midrule
        i3d-Net (Tr-NR, Te-R) & $58.31\%$ & $14.53\%$ & $48.98\%$ \\ 
        i3d-Net (Tr-R, Te-R) & $78.93\%$ & $24.49\%$ & $25.26\%$ \\ 
        i3dLoc (Tr-NR, Te-R) & $83.3\%$ & $52.8\%$ & $39.52\%$\\
        i3dLoc (Tr-R, Te-R) & $87.3\%$ & $58.3\%$ & $42.7\%$\\
        \bottomrule
        \label{table:pr_rot}
        \end{tabular}
    \end{table}
    
    To investigate the generalization of the viewpoint invariant property on different datasets, in the left figure of Fig.~\ref{fig:recall}, we also train/evaluate the performance among different datasets with rotated datasets. 
    Here the \textit{General} dataset is the combination of the other three datasets.
    We can note that when using the trained model from \textit{General} dataset to infer others, i3dLoc can still provide reliable place retrievals. 
    This indicates that our method has the potential to learn place descriptors for large-scale indoor and outdoor datasets at the same time.
    
    We further analyze the average recall by training the networks with/without the rotated datasets.
    Based on the same cross-domain transfer module, we evaluate the performance between i3d-Net and i3dLoc.
    As we can see in Table.~\ref{table:pr_rot}, even when trained with non-rotated datasets, i3dLoc also has reliable average recall in both rotated and non-rotated evaluation.
    It indicates that our method, i3dLoc, has higher generalization ability when training only with limited viewpoints.

    \begin{figure}[t]
    \centering
        \includegraphics[width=\linewidth]{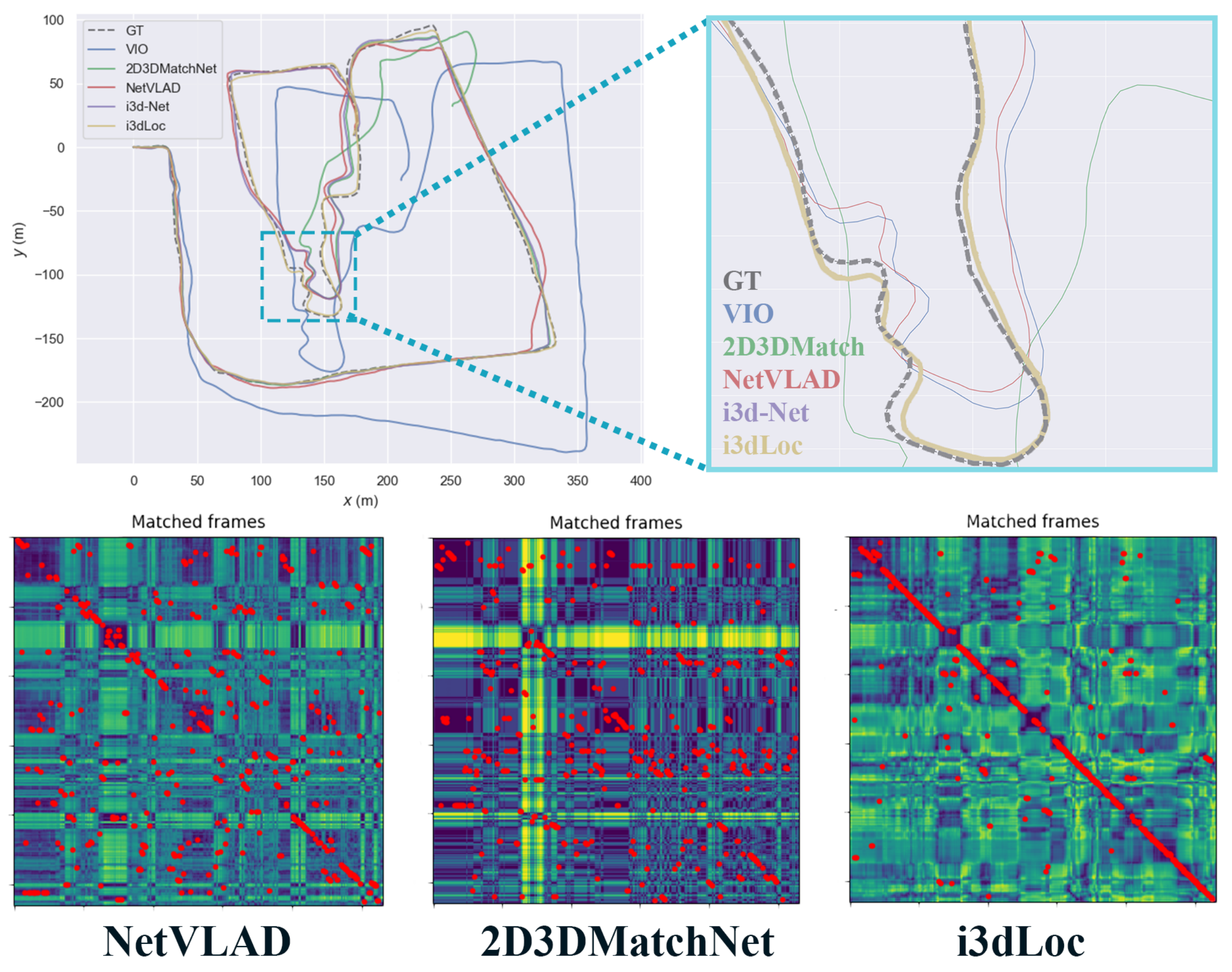}
        \caption{The absolute position error analysis on \textit{Large-scale} dataset.
        The first row shows the localization results on the $2km$ trajectory.
        The second row shows the image-to-range difference matrix, where x- and y-axis represent visual and located range features respectively.
        Red points mean the points with successful place retrieval.}
        \label{fig:accuracy_comp}
    \end{figure}

    \begin{figure*}[ht]
        \centering
        {{\includegraphics[width=0.45\linewidth]{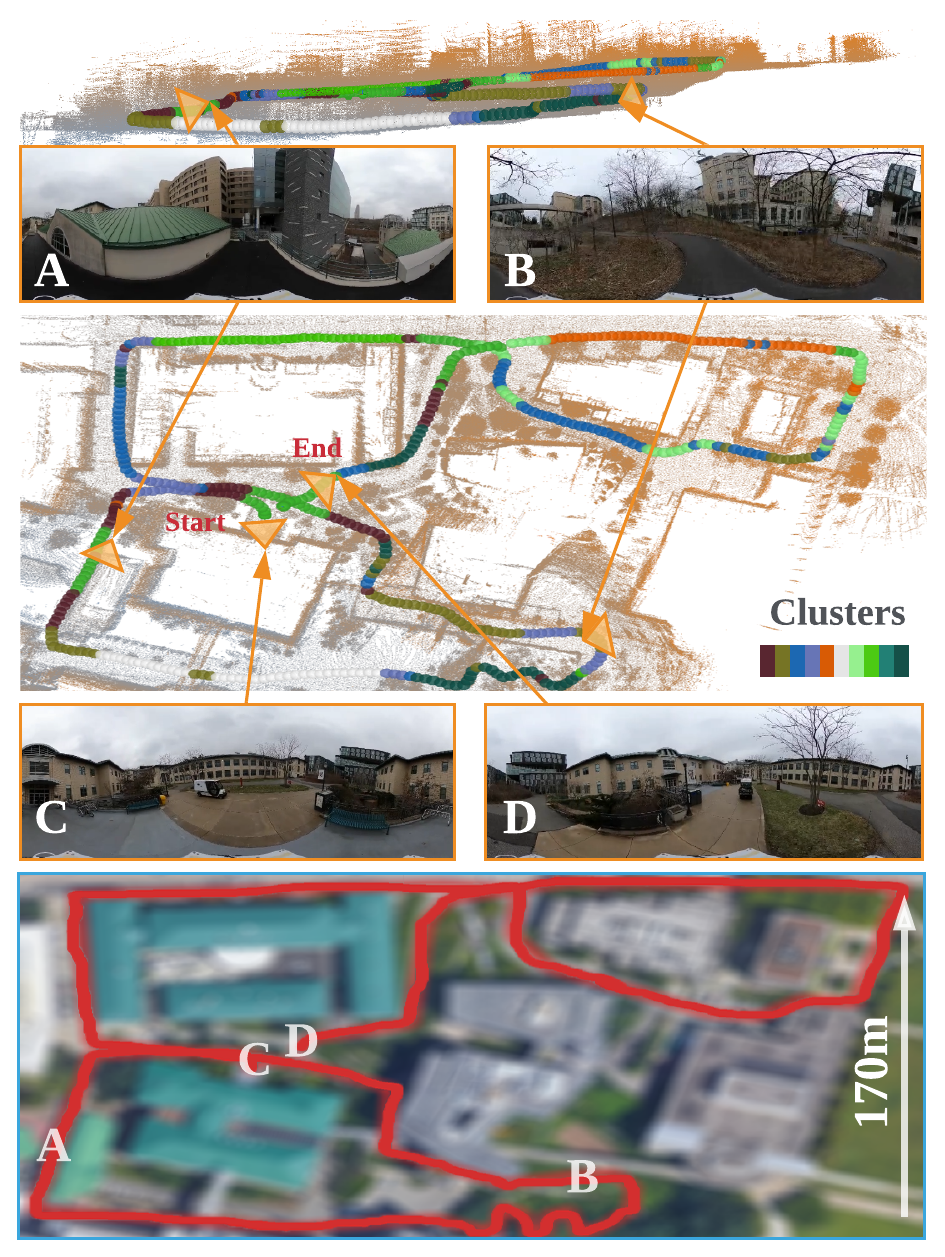} }}
        {{\includegraphics[width=0.45\linewidth]{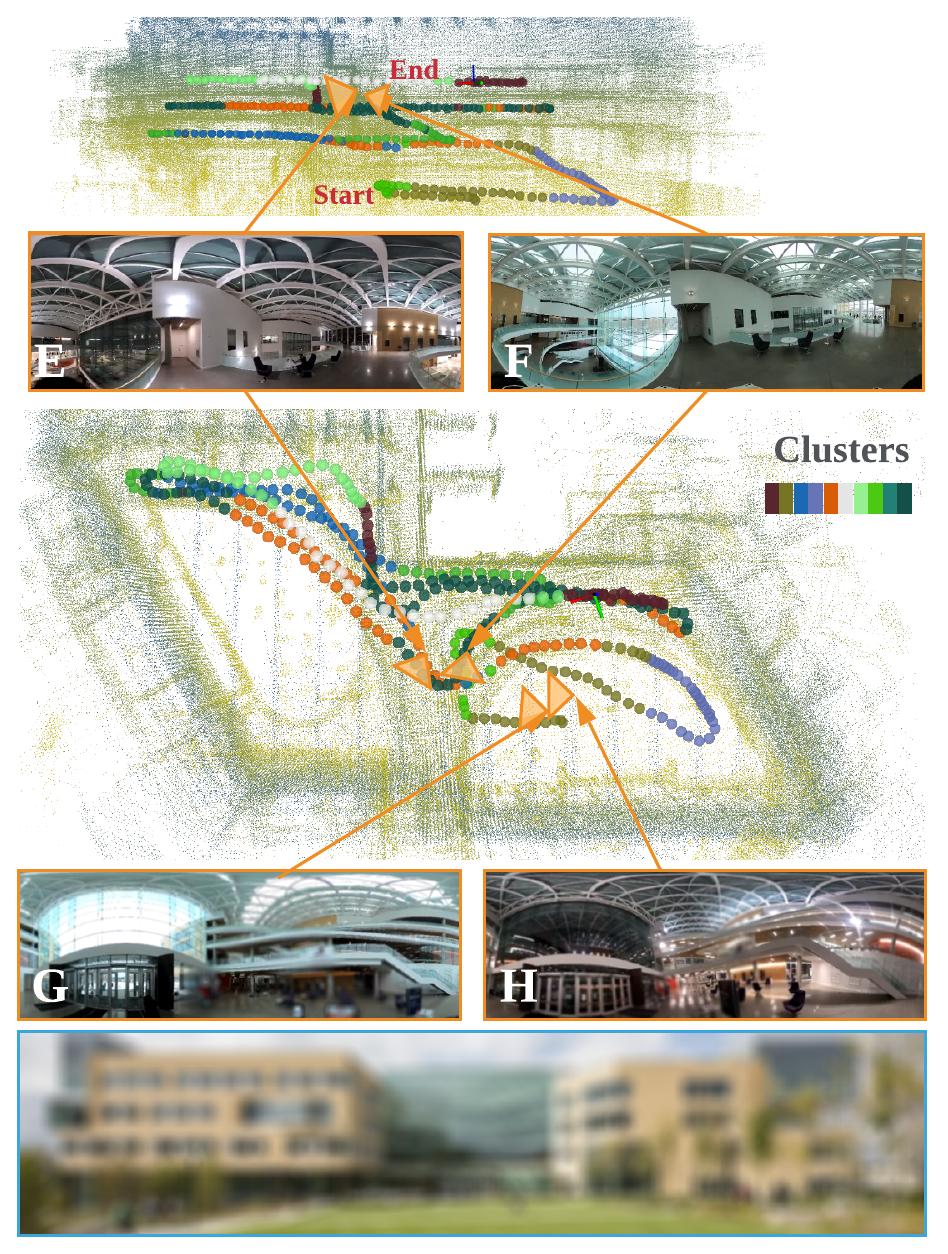} }}%
        \caption{Visual place descriptors in outdoor and indoor environments.
        The top and third rows show a side  and top-down view of the 3d point cloud map. 
        The last rows represent the satellite image and side view of the  outdoor and indoor environments respectively. 
        In each case, We cluster the place descriptors of i3dLoc into $10$ colored groups via K-means method.
        For the outdoor dataset, image $A$ represents a long corridor area, and $B$ represents the unstructured environment in the valley area as we can see in the satellite image.
        Image $C$ and $D$ represent the starting and ending point under different viewpoints.
        For the indoor dataset, images $E,F$ and images $G,H$ represent the ending and starting position under different lighting conditions.}
    \label{fig:place_feature}
    \end{figure*}

\subsection{Online Localization Results}
In this subsection, we further analyze the online localization performance by comparing both traditional visual odometry methods and learning-based approaches.
We conduct this testing on the \textit{Large-scale} dataset, and Fig.~\ref{fig:accuracy_comp} shows the longest trajectory ($2km$) containing both structured and unstructured environments.
The estimated trajectory from ORB-SLAM contains the accumulated odometry drift.
Compared to other learning-based methods, i3dLoc can successfully provide reliable re-localization against the offline 3D point cloud map. 
Other learning-based methods can provide robust place retrieval at the early stage, but fail to find the correspondence in unstructured environments or aggressive viewpoints.
    \begin{table}[t]
        \centering
        \caption{Accuracy \& efficiency in online localization.}
        \begin{tabular}{c c c c c}
        \toprule
        Method & \textit{Mean}(m) & \textit{Std}(m) & \textit{GPU}(MB) & \textit{Time}(ms)\\ \midrule
        \textit{ORB-SLAM~\cite{VPR:ORB-SLAM}} & $48.62$ & $28.65$ & $-$ & $-$ \\ 
        \textit{2D3DMatchNet~\cite{2d3dmatchnet}} & $25.34$ & $36.56$ & $1642$ & $25.4$ \\
        \textit{NetVLAD~\cite{PR:netvlad}} & $12.17$ & $6.37$ & $1478$ & $10.8$ \\
        \textit{i3d-Net} & $7.88$ & $5.78$ & $1121$ & $12.4$ \\ 
        \textit{i3dLoc} & $3.21$ & $2.88$ &  $1273$ & $14.9$ \\ 
        \bottomrule
        \label{table:acc_eff}
        \end{tabular}
    \end{table}
As we can see in the enlarged map on Fig.~\ref{fig:accuracy_comp}, i3dLoc can successfully follow the aggressive trajectory while other methods failed.
Compared to i3dLoc, i3d-Net is more sensitive to local viewpoints differences.
We also plot the cosine feature differences between image queries (x-axis) and the matched range projections (y-axis), where the red points indicates successful retrievals (distance to ground truth within $10m$).
Table.~\ref{table:acc_eff} analyzes the localization accuracy, GPU usage and inferencing time of different methods in online visual localization.
Since the visual localization are combined with the same ORB-SLAM odometry, the localization accuracy of all learning-based methods are better than pure ORB-SLAM.
i3dLoc surpasses all other learning-based methods, while consuming less GPU memory. 
The above property of i3dLoc makes it feasible to run on the embedded system (Nvidia Xavier) for low-cost robots in long-term SLAM and navigation tasks.

\subsection{Discussion}
In Fig.~\ref{fig:place_feature}, we use the same pre-trained model of i3dLoc for online visual localization in both outdoor and Multistory indoor environments.
To demonstrate the similarity among place descriptors, we cluster the extracted descriptors into $10$ classes with each trajectory based on the \textit{K}-means method.
Places sharing with the similar geometric structures are clustered with the same labels.
We can notice that the same areas fall into the same place classes.
i3dLoc has the viewpoint-invariant property, which helps it deal with aggressive viewpoint changes, such as place $B$ in the outdoor environment, and place $E,F$ and a $G,H$ in indoor multistory environments.
However, i3dLoc can only provide coarse localization results, and can not deal with places with continuous 3D geometry structures, i.e., long-corridor and indoor confined spaces.

\section{Conclusions}
\label{sec:conclusions}
    This paper presents a novel image-to-range localization method, i3dLoc, under inconsistent environments.
    The advantage of i3dLoc is that it extracts condition- and viewpoint-invariant features based on our cross-domain transfer learning module and our symmetric feature learning module.
    The experiments on long-term, large-scale, and indoor and outdoor environments demonstrate that our method can surpass both traditional visual SLAM methods and learning-based visual localization methods.
    We also evaluate our method's generalization ability for different environmental conditions and limited viewpoints, which indicates our method can provide reliable place retrieval when trained with both indoor and outdoor environments under variant conditions and viewpoints.
    It leaves us an interesting question: can we enable incremental place feature learning for robotics? 
    In future work, we aim to provide an incremental place feature learning method to enable lifelong visual localization for real-world robots.

\bibliographystyle{unsrtnat}
\bibliography{main}

\end{document}